\documentclass[runningheads]{llncs}

\usepackage{eccv}

\usepackage{eccvabbrv}

\usepackage{graphicx}
\usepackage{booktabs}
\usepackage{amssymb}

\usepackage[accsupp]{axessibility}  

\usepackage{hyperref}

\usepackage{orcidlink}
\usepackage{booktabs}
\usepackage{multirow}

\usepackage{tabularx}
\usepackage[table]{xcolor}
\usepackage{wrapfig}
\definecolor{lightblue}{HTML}{DAEFF9}

\makeatletter
\def\thanks#1{\protected@xdef\@thanks{\@thanks
        \protect\footnotetext{#1}}}
\makeatother
\begin{document}

\title{Looking Back and Forth: Cross-Image Attention Calibration and Attentive Preference Learning for Multi-Image Hallucination Mitigation} 

\titlerunning{Cross-Image Attention Preference Learning for Multi-Image Hallucination}

\author{
Xiaochen Yang\inst{1,2}$^{\dag}$ \and
Hao Fang\inst{1}$^{\dag}$ \and
Jiawei Kong\inst{1} \and
Yaoxin Mao\inst{3} \and \\
Bin Chen\inst{4}$^{\#}$ \and
Shu-Tao Xia\inst{1}
\thanks{$^{\dag}$Equal contribution.}
\thanks{$^{\#}$Corresponding author.}
}

\authorrunning{X. Yang et al.}

\institute{Tsinghua Shenzhen International Graduate School, Tsinghua University \and
Harbin Institute of Technology  \and
Beijing Institute of Technology \and Harbin Institute of Technology, Shenzhen \\ \email{2022211854@stu.hit.edu.cn}, \email{fangh25@mails.tsinghua.edu.cn}}

\maketitle

\begin{abstract}
  Although large vision-language models (LVLMs) have demonstrated remarkable capabilities, they are prone to hallucinations in multi-image tasks. We attribute this issue to limitations in existing attention mechanisms and insufficient cross-image modeling. Inspired by this, we propose a structured hallucination mitigation framework involving \textbf{C}ross-Image \textbf{A}ttention calibration and \textbf{P}reference \textbf{L}earning \textbf{(CAPL)}. CAPL explicitly enhances inter-image interactions at the architectural level while reinforcing reliance on genuine cross-image evidence during training, thereby improving the model’s perception and modeling of cross-image associations. Specifically, we (i) introduce a selectable image token interaction attention mechanism to establish fine-grained cross-image entity alignment and information flow; (ii) design a cross-image modeling–based preference optimization strategy that contrasts reasoning outcomes under full inter-image interaction and those obtained when images are mutually invisible, encouraging the model to ground its predictions in authentic visual evidence and mitigating erroneous inferences driven by textual priors. Experimental results demonstrate that CAPL consistently improves performance across multiple model architectures, achieving stable gains on both multi-image hallucination and general benchmarks. Notably, performance on single-image visual tasks remains stable or slightly improves, indicating strong generalization capability.
\end{abstract}

\section{Introduction}
\label{sec:intro}

Recently, large-scale visual language models (LVLMs) have achieved significant progress in single-image visual question answering, description generation, and reasoning tasks. With the growing demand for multi-image inputs in real-world applications, such as multi-view comparison and cross-image information integration, models are required not only to understand individual images but also to establish relationships and consistency across images. Accordingly, models like Idefics\cite{laurenccon2024matters} and Qwen-VL\cite{wang2024qwen2} have begun to support multi-image inputs and demonstrate capabilities in multi-image understanding tasks. However, LVLMs\cite{liu2023visual} exhibit serious hallucination issues in downstream tasks \cite{liu2024survey}, \ie, generate plausible yet completely factually incorrect answers. Most existing mitigation methods focus on single-image scenarios and reduce hallucinations through decoding stratgies\cite{huang2024opera,fang2025grounding} or alignment training \cite{jiang2024hallucination}. While several decoding-based studies \cite{tian2025identifying,li2025mihbench} have also been proposed to handle the multi-image hallucination, their improvements remain limited since they only adjust the decoding distribution via local structural modifications, without fundamentally enhancing the model’s cross-image interactions. Another line of research \cite{liu2024mia,zhang2025perl} incorporates delicate training to enhance multi-image reasoning capability for hallucination reduction. However, they still treat each image as an independent context without explicitly modeling inter-image relationships, making it difficult to effectively address hallucinations in multi-image tasks.

Multi-image tasks require the model not only to understand the individual semantics of each image but also to establish symmetric and stable relational modeling across images. Existing Transformer-based autoregressive LVLMs process text and image tokens under a unified causal attention framework \cite{yang2021causal}. In this setup, multi-image inputs are processed sequentially, \ie, later images can attend to earlier ones, while earlier images have no access to representations of later images, thereby introducing an inherent position bias. 
This asymmetric information flow undermines semantic equivalence in cross-image relational modeling, making it difficult for the model to establish stable associations across images.
\begin{figure}[tb]
  \centering
  \includegraphics[width=\linewidth]{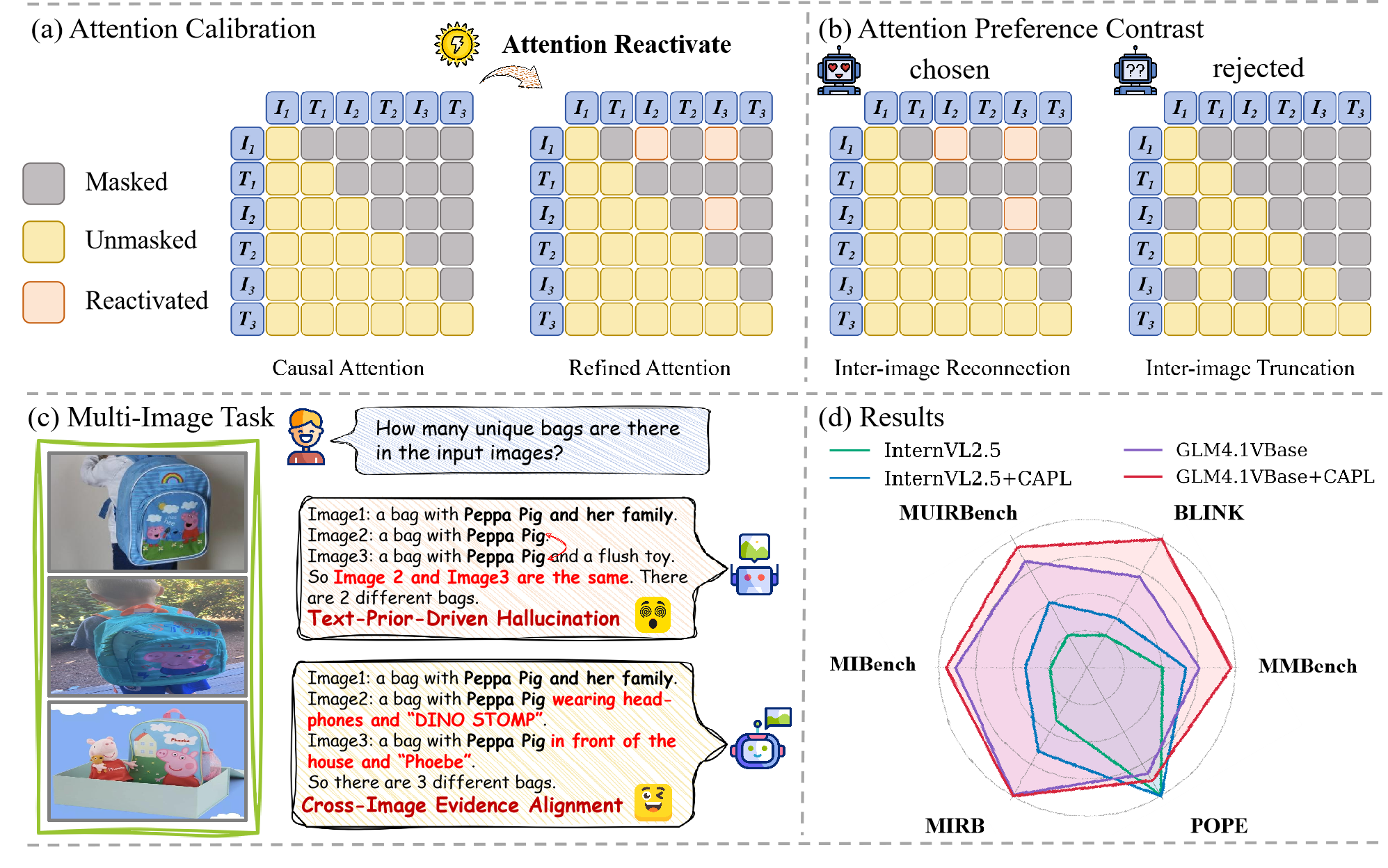}
  \caption{Overview of our training framework and evaluation. 
    (a) Attention Calibration: Reactivates cross-image token attention on top of causal attention. 
    (b) Attentive Preference Contrast: Generates positive and negative attention for preference learning via inter-image reconnection and truncation. 
    (c) Multi-Image Task: Illustrates a consistency question; incorrect answers ignore inter-image distinctions while correct ones leverage key inter-image information. 
    (d) Evaluation: Our framework perform well on multi-image hallucination and general tasks, while preserving single-image capabilities.}
  \label{fig:motivation}
\end{figure}
Under such an order-biased interaction pattern, the cross-image information flow exhibits a predominantly unidirectional propagation, which hinders the model from establishing symmetric and stable relational alignment at the image-token level. As a result, multi-image reasoning may degenerate into superficial correlation matching over text tokens, rather than genuine relational modeling grounded in visual ones. As illustrated by the first response in Fig.~\ref{fig:motivation}(c), model predictions in such cases become increasingly susceptible to language priors\cite{ghosh2024visual} and autoregressive generation biases due to insufficient cross-image visual interactions, ultimately leading to unreliable relational inferences.

To address the aforementioned issues, we propose a novel mitigation framework comprising of \textbf{C}ross-Image \textbf{A}ttention and \textbf{P}reference \textbf{L}earning \textbf{(CAPL)}. Specifically, we first introduce a selective cross-image token mutual attention mechanism to reduce positional bias, as shown in Fig.~\ref{fig:motivation}(a). This strategy enables key tokens from different images to interact with each other explicitly during inference. By selectively reactivating the attention between critical cross-image token pairs, CAPL facilitates structured inter-image information exchange and promotes a more balanced bidirectional attention flow, which effectively alleviates relational modeling limitations induced by the unidirectional causal attention mechanism. 
Next, we note that LVLMs are generally pretrained under the causal attention paradigm and may lack adaptability to the proposed bidirectional cross-image attention. As a result, directly adapting them to this design may not fully unleash the potential benefits. To handle this challenge, we propose to construct dedicated samples to conduct further training with the proposed attention technique, which enables better adaptation to this mutual interaction paradigm. An intuitive training approach is to perform supervised fine-tuning (SFT) \cite{bai2025hallucination} on the LVLM using high-quality data samples. However, we reflect that the SFT objective utilizes only positive samples, without considering and penalizing the models' inherent hallucinated responses, which may fail to effectively suppress their own erroneous reasoning patterns. 

Based on these analyses, we introduce an attentive preference alignment approach based on Direct Preference Optimization (DPO)\cite{rafailov2023direct}. 
To obtain preferred non-hallucinated samples, we equip the LVLMs with the proposed attention correction mechanism and query them with different questions. The generated responses are further improved by incorporating feedback from an advanced Qwen3 model.
For rejected sample construction, we aim to encourage the LVLM to fully expose its hallucination behaviors. Concretely, we draw inspiration from the causal attention mechanism in multi-image scenarios from an inverse perspective. In the original causal attention, information flows unidirectionally from earlier to later images. To generate more hallucinated negative responses, we exploit this limitation by completely truncating all cross-image attention connections, enforcing representational independence across all images.
As a result, the model is forced to rely solely on individual images and language priors during inference, increasing the likelihood of hallucinated responses. As illustrated in Fig.~\ref{fig:motivation}(b), by constructing preferred–rejected response pairs under this controlled setting, we guide the model to overcome its inherent hallucination behaviors and mitigate the generation of erroneous reasoning trajectories.

In summary, our main contributions are as follows:
\begin{itemize}
    \item We analyze the structural causes of hallucination in multi-image reasoning, identifying imbalanced visual information flow and insufficient cross-image semantic association as key factors that limit multi-image reasoning performance.
    \item We propose a novel framework CAPL, which integrates selective cross-image attention with preference alignment training, enhancing semantic interaction among critical cross-image tokens and reinforce the model to better perceive and utilize inter-image interactions.
    \item Extensive experiments(Fig.~\ref{fig:motivation}(d)) demonstrate that our method generalizes well across multiple recent vision-language models, significantly reducing hallucination and improving reasoning performance on multi-image tasks.
\end{itemize}

\section{Related Work}

\subsection{Large Vision-Language Models}
Recent years have witnessed significant progress in Large Vision-Language Models across a wide range of multimodal understanding and generation tasks. Representative approaches typically adopt a two-stage framework that aligns a visual encoder with a large language model. For example, the LLaVA series \cite{liu2023visual} maps visual features into the language space via a projection layer followed by instruction tuning; Qwen-VL\cite{bai2025qwen3} and InternVL\cite{chen2024internvl} further enhance high-resolution visual perception and support multi-image inputs; GLM-4V\cite{hong2024cogvlm2} enhances visual understanding while maintaining strong language reasoning capabilities. These methods generally follow a unified architecture of “visual encoder + linear projection + LLM”, and rely on instruction-tuning data to achieve multimodal alignment, enabling applications such as multimodal dialogue \cite{liu2024mmdu}, visual question answering (VQA) \cite{antol2015vqa} and image captioning \cite{rohrbach2018object}. However, in multi-image scenarios, existing models typically adopt relatively simple strategies, such as concatenating visual tokens or directly fusing shared representations, without explicitly modeling structured relationships across images. This limitation poses significant challenges for multi-image reasoning and consistency modeling.

\subsection{Multimodal Hallucination in LVLMs}
Although LVLMs achieve strong performance across various benchmarks, multimodal hallucination remains a persistent challenge, which can be generally categorized to single-image hallucination and multi-image hallucination.

\textbf{Single-image hallucination.} This issue typically stems from data bias\cite{hu2023ciem}, annotation noise\cite{liu2023mitigating}, insufficient vision–language alignment\cite{sun2024aligning,liu2024improved}, and the model’s over-reliance on language priors\cite{ghosh2024visual}. To mitigate this problem, existing approaches focus on several directions, including data refinement, bias correction\cite{hu2023ciem}, improved cross-modal alignment architectures\cite{jiang2024hallucination}, and the introduction of preference alignment\cite{zhao2023beyond, fang2026seeing} or decoding constraints\cite{leng2024mitigating, fang2025grounding} to improve consistency between generated content and visual evidence. In addition, some studies\cite{favero2024multi} analyze hallucinations from the perspective of attention distribution and token grounding, revealing that models may attend to irrelevant visual regions during generation. Other works\cite{dai2023plausible} attempt to strengthen visual grounding by incorporating external detectors or region-level supervision. These methods have shown effectiveness in reducing hallucinations in single-image scenarios, but their applicability to more complex multi-image settings remains limited.

\textbf{Multi-image hallucination}. With the emergence of multi-image tasks, recent work has begun to investigate error patterns in multi-image scenarios, such as incorrectly merging information from different images during comparison or introducing nonexistent entities and attributes in relational reasoning. Some training-free methods \cite{li2025mihbench} attempt to alleviate bias toward particular images by averaging attention across images. However, their effectiveness remains limited since the model parameters are not updated. Preference learning approaches \cite{liu2024mia,li2025zooming} construct preference pairs based on attention distributions over key images or from global and targeted perspectives. Nevertheless, these works mainly address quasi single-image settings within multi-image inputs, where images are often semantically unrelated, resulting in limited attention to modeling genuine cross-image semantic relationships. Furthermore, the complex dependencies and fine-grained differences between images in multi-image tasks make hallucination problems more difficult to mitigate. Especially when inter-image information flow is inadequately modeled, models are more prone to semantic confusion and erroneous reasoning across images.

\section{Method}
\subsection{Preliminary}
In the multi-image setting, we assume that the input consists of $N$ images and their corresponding textual descriptions. 
The $k$-th image is encoded by a visual encoder into a sequence of visual tokens:

\begin{equation}
\mathbf{I}_k = [v_{k,1}, \dots, v_{k,\tau_k}], 
\quad k = 1, \dots, N
\end{equation}

\noindent where $v_{k,j}$ denotes the $j$-th visual token of the $k$-th image, 
and $\tau_k$ represents the length of the visual token sequence for that image.
Similarly, the associated text is tokenized into a sequence of textual tokens:

\begin{equation}
\mathbf{T}_k = [t_{k,1}, \dots, t_{k,L_k}], 
\quad k = 1, \dots, N
\end{equation}

\noindent where $t_{k,j}$ denotes the $j$-th token of the $k$-th text segment, 
and $L_k$ denotes the length of the corresponding text token sequence.
Following common LVLM input formatting, the multimodal input is organized according to its original ordering in an interleaved image-text format to form a unified sequence.

\begin{equation}
\mathbf{z} =
[\mathbf{I}_1, \mathbf{T}_1,
 \mathbf{I}_2, \mathbf{T}_2,
 \dots,
 \mathbf{I}_N, \mathbf{T}_N].
\end{equation}

Let $T$ denote the total length of the concatenated sequence. 
Under the autoregressive Transformer framework, self-attention is computed as:

\begin{equation}
\mathbf{A} =
\mathrm{Softmax}
\left(
\frac{\mathbf{Q}\mathbf{K}^\top}{\sqrt{d}} + \mathbf{M}
\right),
\end{equation}

\noindent where $\mathbf{Q}, \mathbf{K}, \mathbf{V} \in \mathbb{R}^{T \times d}$ 
are the query, key, and value matrices obtained through linear projections, 
and $\mathbf{M} \in \mathbb{R}^{T \times T}$ is the attention mask matrix.
Under the standard causal attention mechanism, $\mathbf{M}$ is defined as:

\begin{equation}
\mathbf{M}_{ij} =
\begin{cases}
0, & j \le i, \\
-\infty, & j > i,
\end{cases}
\end{equation}

\noindent which constrains each token to attend only to itself and preceding tokens, 
thereby ensuring autoregressive generation.

Since both visual and textual tokens are embedded into a unified sequence space, 
cross-image and cross-modal dependencies are implicitly determined by the relative 
positions of tokens in the sequence and the causal attention mask.

\begin{figure}[tb]
  \centering
  \includegraphics[width=\linewidth]{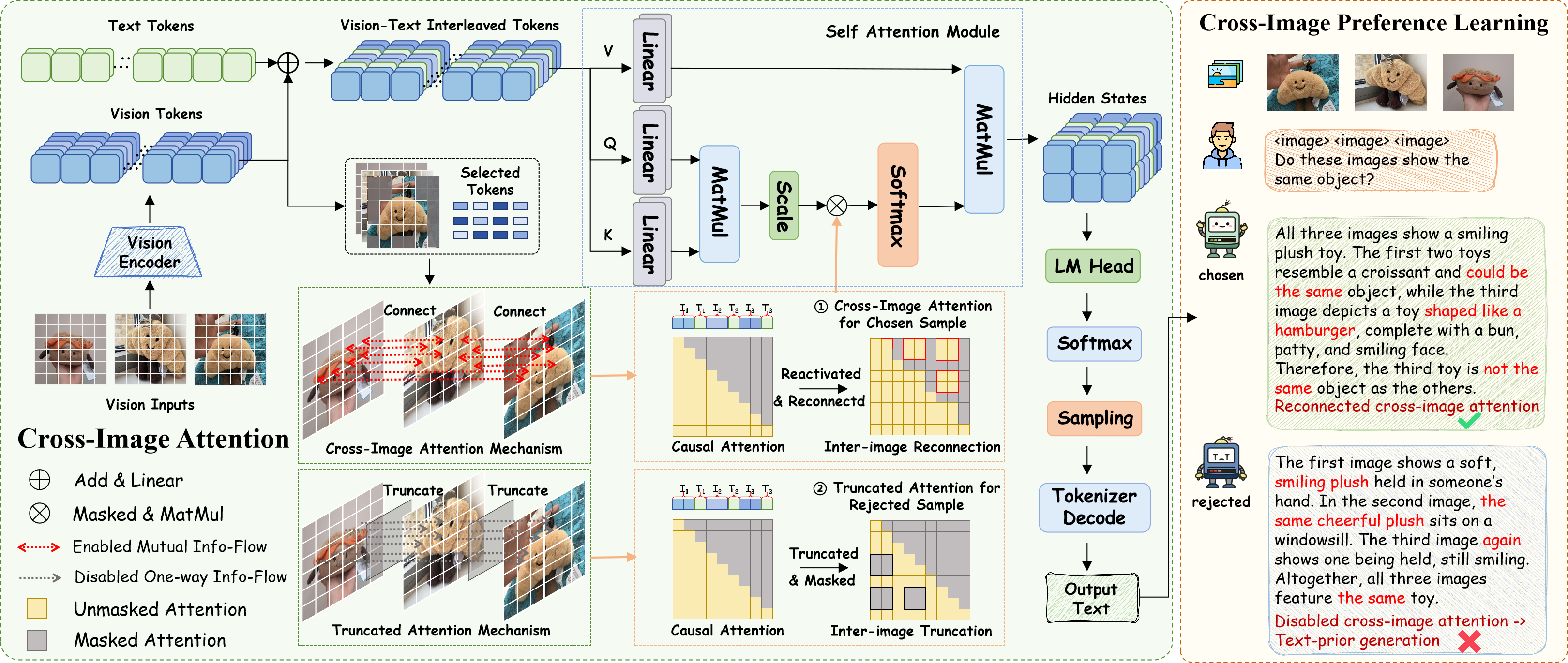}
  \caption{The overview of the CAPL framework. The pipeline consists of attention modification and preference learning. Image tokens are ranked to select key tokens, and the original causal mask is modified into enhanced cross-image attention (reactivating inter-image token interactions) and truncated cross-image attention (blocking inter-image interactions). The enhanced mask is used for inference and positive sample construction, while the truncated mask generates negative samples for DPO training.
  }
  \label{fig:pipeline}
\end{figure}

\subsection{Selective Cross-Image Token Interaction}
In the multi-image autoregressive modeling framework, the standard causal attention induces a unidirectional cross-image information flow: tokens from later images can attend to representations of earlier images, whereas tokens from earlier images cannot access information from later ones. This asymmetry introduces an explicit ordering bias and weakens the cross-image symmetry required for modeling mutual semantic relationships. To alleviate this limitation, we propose a cross-image attention interaction mechanism that activates attention connections between tokens from different images. By breaking the unidirectional constraint, this design allows earlier images to perceive later ones and enables bidirectional information flow across visual input. We implement this by removing the causal mask between different images while preserving that within each image to retain intra-image positional structure and order bias.

Formally, let $g(i)$ denote the index of the image to which token $i$ belongs. The cross-image mask $\mathbf{M}^{\mathrm{cross}}$ is defined as:

\begin{equation}
\mathbf{M}^{\mathrm{cross}}_{ij} =
\begin{cases}
0, & g(i) \neq g(j), \\
\mathbf{M}_{ij}^{\mathrm{causal}}, & g(i) = g(j),
\end{cases}
\label{eq:cross_mask}
\end{equation}

\noindent which allows bidirectional attention only between tokens from different images, while retaining the original causal attention structure within each image.

Since information density and semantic relevance may vary across images, enabling full cross-image attention may introduce redundant interactions. To address this issue, we further introduce a key-token selection mechanism based on embedding energy. Let the visual tokens of the $k$-th image be denoted as $\mathbf{h}_{k,1}, \dots, \mathbf{h}_{k,\tau_k} \in \mathbb{R}^d$, and define their response intensity as:

\begin{equation}
s_{k,i} = \|\mathbf{h}_{k,i}\|_2.
\label{eq:energy}
\end{equation}

We introduce a ratio parameter $\rho \in (0,1]$ to control the selection range of key tokens. For each image, the top $\lfloor \rho \tau_k \rfloor$ tokens with the highest response intensity are selected to form the key-token set:

\begin{equation}
\mathcal{S}_k = \{ i \mid s_{k,i} \text{ ranks among the top } \rho \tau_k \}.
\label{eq:selection}
\end{equation}

Based on the cross-image mask defined in Eq.~\ref{eq:cross_mask}, we further construct the selective cross-image mask $\mathbf{M}^{\mathrm{cross\_sel}}$ as:

\begin{equation}
\mathbf{M}^{\mathrm{cross\_sel}}_{ij} =
\begin{cases}
0, & g(i) \neq g(j), \ i \in \mathcal{S}_{g(i)}, \ j \in \mathcal{S}_{g(j)}, \\
\mathbf{M}^{\mathrm{causal}}_{ij}, & \text{otherwise}.
\end{cases}
\label{eq:cross_sel_mask}
\end{equation}

This mechanism enables effective interactions only between key tokens from different images, thereby enhancing cross-image semantic association modeling while suppressing irrelevant information propagation.

The proposed interaction attention removes the cross-image causal order bias, enabling symmetric visibility across images and facilitating relational reasoning. However, some multi-image tasks involve temporal dependencies or single-image queries, where preserving the sequential structure of images remains necessary. Therefore, completely replacing causal attention with cross-image attention may not be optimal. To balance these requirements, we fuse the selective cross-image attention with the original causal attention and compute the final attention weights as their equal-weight combination:

\begin{equation}
\mathbf{A}^{\mathrm{fuse}} =
\frac{1}{2}
\left(
\mathbf{A}^{\mathrm{causal}} +
\mathbf{A}^{\mathrm{cross\_sel}}
\right).
\label{eq:final_attention}
\end{equation}

Furthermore, to prevent cross-image interactions from disrupting the original autoregressive modeling pathway, we adopt an alternating mask strategy at the decoder-layer level: odd-numbered layers apply the selective cross-image mask $\mathbf{M}^{\mathrm{cross\_sel}}$, while even-numbered layers retain the original causal mask $\mathbf{M}^{\mathrm{causal}}$. This hierarchical alternating design enhances cross-image relational modeling while preserving stability and generalization in non-cross-image tasks.

\subsection{Cross-Image Attention Guided DPO for Preference Learning}
Most vision-language models are pre-trained using an autoregressive causal attention paradigm, so merely modifying the attention mechanism during inference  yields only temporary improvements. Traditional SFT can force the model to mimic positive samples but cannot suppress its inherent hallucination tendencies. Existing approaches such as RLHF \cite{sun2024aligning} and RLAIF \cite{lee2023rlaif} require costly construction of negative samples and often struggle to capture the model's internal response patterns and latent hallucination directions, thus failing to effectively generate hallucination-targeted negatives. In contrast, Direct Preference Optimization (DPO) offers a more flexible training paradigm. By explicitly contrasting the generation probabilities of positive and negative samples, it guides the model to prefer reliable cross-image outputs while suppressing potential hallucination pathways. This approach not only reshapes the output distribution, but also promotes stable cross-image relational modeling in the parameter space, rather than relying solely on inference-time adjustments.

The construction of positive and negative samples plays a crucial role in the DPO framework. For positive samples, we generate outputs using the selective cross-image attention mechanism and further refine them with Qwen3 to ensure correctness. To effectively guide the model to recognize and correct its hallucination behaviors, we propose to construct negative samples that expose its inherent erroneous reasoning. 
Directly using outputs from the model as negatives is of limited effectiveness, as these outputs often fail to reveal its hallucination patterns and therefore provide weak optimization signals. Specifically, we draw inspiration from the causal attention mechanism in multi-image settings from a complementary perspective. While standard causal attention allows information to flow unidirectionally between images, we exploit this property to induce hallucinations by truncating attention between tokens from different images, enforcing representational independence across all images. This forces the model to rely solely on individual images when generating negative samples, while reasoning about cross-image relationships degenerates to inference based on text priors. Formally, we define the truncated attention mask $\mathbf{M}_{\text{trunc}}$ as:
\begin{equation}
\mathbf{M}_{ij}^{\text{trunc}} =
\begin{cases}
\mathbf{M}_{ij}^{\text{causal}}, & g(i) = g(j), \\
-\infty, & g(i) \neq g(j),
\end{cases}
\label{eq:trunc_mask}
\end{equation}
where $g(i)$ denotes the index of the image to which token $i$ belongs. Text generated under this mask serves as the negative sample.

Given a $x_\text{I,T}$ with both input images and the text question, let $y^+$ and $y^-$ denote the outputs generated under cross-image attention and the truncated attention mask, respectively. The DPO objective is:
\begin{equation}
\mathcal{L}_{\text{DPO}}(\pi_\theta; \pi_\text{ref}) = 
-\log \sigma \Big(
\beta \, \log \frac{\pi_\theta(y^+ \mid x_\text{I,T})}{\pi_\text{ref}(y^+ \mid x_\text{I,T})}
-
\beta \, \log \frac{\pi_\theta(y^- \mid x_\text{I,T})}{\pi_\text{ref}(y^- \mid x_\text{I,T})}
\Big) ,
\label{eq:dpo_loss}
\end{equation}
where $\pi_\theta(y \mid x_\text{I,T})$ is the probability of generating $y$ given the image–text input $x_\text{I,T}$, and $\pi_\text{ref}(y \mid x_\text{I,T})$ is the reference model probability. Through this training process, the model learns to explicitly prefer outputs produced under cross-image bidirectional attention. As a result, the utilization of cross-image information is gradually strengthened in the parameter space, leading to improved semantic relational modeling across images and effective suppression of multi-image hallucinations. This contrastive training strategy further enables the model to robustly handle complex multi-image inputs while preserving its original language generation capability.

\subsection{Training Optimization}

While DPO optimizes the relative preference between positive and negative responses, it does not explicitly enforce the model to imitate the token-level generation trajectory of high-quality positive samples. As a result, the model may learn preference ordering without fully internalizing the structured cross-image reasoning process. To address this limitation, we incorporate a negative log-likelihood (NLL) loss on the positive samples. Given input $x_\text{I,T}$ and its corresponding positive response $y^+ = \{y^+_1, \dots, y^+_T\}$, the NLL objective is defined as:

\begin{equation}
\mathcal{L}_{\text{NLL}}(\pi_\theta) = - \sum_{t=1}^{T} \log \pi_\theta \big( y^+_t \mid x_\text{I,T}, y^+_{<t} \big),
\label{eq:nll_loss}
\end{equation}

\noindent where $\pi_\theta$ denotes the model policy. This objective encourages the model to follow generation trajectories aligned with cross-image semantic reasoning.

Finally, the overall training objective combines DPO and NLL as:

\begin{equation}
\mathcal{L}_{\text{total}} = 
\mathcal{L}_{\text{DPO}}(\pi_\theta; \pi_\text{ref}) + \lambda \mathcal{L}_{\text{NLL}}(\pi_\theta),
\label{eq:total_loss}
\end{equation}

\noindent where $\lambda$ is a balancing coefficient. The DPO term enforces relative preference alignment between contrasted samples, while the NLL term reinforces token-level imitation of reliable cross-image reasoning paths. This hybrid objective improves optimization stability and further enhances cross-image semantic consistency.

\section{Experiments}
\label{sec:experiment}
\subsection{Experimental Setup}
\textbf{Baselines.}
We apply our method to three different base models: Qwen2.5-VL-7B-Instruct\cite{bai2025qwen3}, InternVL2.5-8B\cite{chen2024expanding}, and GLM4.1VBase-9B\cite{hong2025glm}. 
As baselines, we use the original pre-trained models without additional alignment. 
For a broader comparison, we report results of other LVLMs, including Idefics2\cite{laurenccon2024matters}, Idefics3\cite{laurenccon2025foundation}, LLaVA-OV\cite{li2024llava}, LLaVA-Next\cite{li2024llava}, InternVL2\cite{chen2024internvl} and Qwen2VL\cite{wang2024qwen2}. 

\textbf{Benchmarks.} 
To comprehensively evaluate the effectiveness and generalization ability of our method in multi-image scenarios, we conduct experiments on diverse benchmarks. We first evaluate on BLINK\cite{fu2024blink} and MUIRBench\cite{wang2024muirbench}, two representative benchmarks specifically designed to evaluate \textbf{multi-image hallucination}, which measure the model’s ability to establish semantic associations across images and suppress hallucinated reasoning. We further evaluate on mainstream \textbf{multi-image general capability} evaluation benchmarks, including NLVR2\cite{suhr2019corpus}, QBench2\cite{zhang2024q}, MIBench\cite{liu2024mibench}, and MIRB\cite{zhao2024benchmarking}, which primarily focus on overall general understanding and reasoning rather than explicitly targeting hallucination. In addition, we conduct experiments on \textbf{single-image} benchmarks such as POPE\cite{li2023evaluating}, CHAIR \cite{rohrbach2018object}, MMBench\cite{liu2024mmbench} and AMBER\cite{wang2023amber} to examine robustness under cross-domain settings. Through evaluation on these diverse benchmarks, we aim to demonstrate the effectiveness and robustness of our approach across different task types and domain scenarios.

\textbf{Implementation Details}
We construct a training set of 3.6K samples collected from WikiArt, MDVP\cite{lin2024draw}, Mantis-Instruct\cite{jiang2024mantis}, and VLM2Bench\cite{zhang2025vlm2}, ensuring that no data overlaps with the evaluation benchmarks. Considering the architectural differences among base models, we adopt model-specific hyperparameters. For the top cross-image token selection ratio $\rho$, the values for Qwen2.5-VL, InternVL2.5, and GLM4.1VBase are set to 0.95, 0.95, and 0.9, respectively. For the NLL loss weight $\lambda$, the corresponding values are 2.5, 2.5, and 2. During training, we use a learning rate of $1e{-4}$ and apply LoRA fine-tuning with a rank of 16. Regarding training frameworks, Qwen2.5-VL and GLM4.1VBase are trained with LLaMAFactory\cite{zheng2024llamafactory}, while InternVL2.5 is trained with MS-Swift\cite{zhao2025swift}. All models are evaluated on VLMEvalKit\cite{duan2024vlmevalkit} to ensure consistency and comparability, and all experiments are conducted on NVIDIA L20 GPUs.

\begin{table*}[t]
\centering
\small
\setlength{\tabcolsep}{4pt}

\caption{
Performance comparison on hallucination and general capability evaluation in multi-image scenarios. BLINK and MUIRBench are hallucination tasks, which constitute the core evaluation of our method. “–” indicates results that cannot be measured.
}
\label{tab:main_results}

\begin{tabularx}{\textwidth}{l c >{\centering\arraybackslash}m{1.25cm} >{\centering\arraybackslash}m{1.25cm} *{4}{>{\centering\arraybackslash}X}}
\toprule

\multirow{2}{*}{Models} & 
\multirow{2}{*}{Params} &
\multicolumn{2}{c}{Hallucination Eval} &
\multicolumn{4}{c}{General Capability Eval} \\

\cmidrule(lr){3-4}
\cmidrule(lr){5-8}

& & BLINK & MUIRB & NLVR2 & QBench2 & MIBench & MIRB \\

\midrule

Idefics2      & 8B & 45.24 & 29.85 & 56.81 & 38.6 & 49.28 & 37.56 \\
Idefics3      & 8B & 48.34 & 30.96 & 85.14 & 64.2 & 54.18 & -- \\
LLaVA-OV      & 7B & 44.77 & 30.85 & 86.82 & 70.1 & 62.37 & 47.3 \\
LLaVA-Next    & 7B & 41.19 & 30.50  & 50.34 & 39.5 & --    & -- \\
InternVL2     & 7B & 50.34 & 45.61 & 77.68 & 69.8 & 59.37 & 53.15 \\
Qwen2VL       & 7B & 53.17 & 39.57 & 87.41 & 76.8 & 69.20 & 31.68 \\

\midrule
Qwen2.5-VL     & 7B & 54.60 & 58.42 & 79.85 & 71.1 & 72.42 & 52.73 \\
\rowcolor{lightblue}
+CAPL (Ours)  & 7B & \textbf{57.76} & \textbf{62.00} & 80.05 & 72.4 & 71.06 & 56.55 \\

\midrule
InternVL2.5   & 8B & 54.81 & 48.54 & 90.42 & 75.5 & 63.42 & 54.18 \\
\rowcolor{lightblue}
+CAPL (Ours)  & 8B & \textbf{55.76} & \textbf{52.12} & 90.13 & 75.3 & 65.11 & 56.55 \\

\midrule
GLM4.1VBase   & 9B & 58.17 & 57.84 & 84.98 & 74.4 & 70.86 & 59.96 \\
\rowcolor{lightblue}
+CAPL (Ours)  & 9B & \textbf{61.33} & \textbf{60.57} & 84.87 & 73.6 & 71.70 & 60.06 \\

\bottomrule
\end{tabularx}

\end{table*}

\subsection{Main Results}
\textbf{Results on Multi-Image Hallucination Tasks.}
Results on the multi-image hallucination benchmarks BLINK and MUIRBench are shown in Tab.~\ref{tab:main_results}. Overall, CAPL brings consistent improvements across Qwen2.5-VL, InternVL2.5, and GLM4.1VBase, demonstrating strong cross-architecture generalization. On BLINK, all models achieve gains of approximately 1--3 points. On the large-scale MUIRBench benchmark, which emphasizes complex cross-image relational reasoning, the improvements are more pronounced, reaching over 3.5 points in the best case. Notably, even for the strong baseline GLM4.1VBase-9B, incorporating CAPL still yields steady improvements. This observation suggests that even advanced models rely on relatively naive attention mechanisms in multi-image settings, leaving room for improving inter-image interaction structures. Overall, explicitly modeling cross-image attention together with preference optimization consistently enhances the reliability of multi-image reasoning and effectively mitigates hallucinations caused by weak cross-image associations.

\textbf{Results on Multi-Image General Tasks.}
We further assess CAPL on general multi-image general benchmarks in Tab.~\ref{tab:main_results}, including NLVR2, QBench2, MIBench, and MIRB. These tasks evaluate whether the model can determine the correctness of descriptions given multiple images, emphasizing multi-image reasoning and general capability rather than hallucination detection. Although CAPL is primarily designed to mitigate hallucinations in multi-image tasks, we observe that its performance remains stable or even shows slight improvements across most multi-image general benchmarks. We attribute this gain to explicit cross-image attention modeling, which regulates information flow between images and increases the effective contribution of visual tokens during reasoning, thereby strengthening the model’s reliance on visual evidence.

\begin{table*}[t]
\centering
\small
\setlength{\tabcolsep}{4pt}

\caption{
Performance comparison on single-image benchmarks.
$\uparrow$ indicates higher is better; $\downarrow$ indicates lower is better.
}
\label{tab:single_results}
\begin{tabularx}{\textwidth}{l *{5}{>{\centering\arraybackslash}X}}
\toprule
Models 
& POPE $\uparrow$ 
& CHAIRi $\downarrow$ 
& CHAIRs $\downarrow$ 
& MMB $\uparrow$ 
& AMBER $\uparrow$ \\

\midrule

Idefics2   & 86.35 & 4.8 & 7.6  & 76.46 & 86.40 \\
Idefics3   & 85.50  & 7.6 & 40.2 & 75.51 & 84.09 \\
LLaVA-OV   & 89.15 & 8.7 & 34.0   & 81.70  & 83.99 \\
LLaVA-Next & 87.42 & 13.3 & 22.6 & 69.93 & 84.96 \\
InternVL2  & 84.39 & 8.3 & 34.0 & 82.90 & 86.00 \\
Qwen2VL    & 84.00 & 10.5 & 28.8 & 81.53 & 85.96 \\
\midrule
Qwen2.5-VL     & 81.23 & \textbf{7.2} & 29.6 & \textbf{88.15} & 85.13 \\

\rowcolor{lightblue}
+CAPL (Ours) & \textbf{82.94} & 7.5 & \textbf{28.6} & 87.48 & \textbf{85.25} \\

\midrule
InternVL2.5    & 88.94 & \textbf{6.4} & 24.4 & 84.11 & 88.99 \\
\rowcolor{lightblue}
+CAPL (Ours) & \textbf{89.08} & 7.3 & \textbf{23.8} & \textbf{84.27} & \textbf{89.79} \\

\midrule
GLM4.1VBase       & 84.79 & 7.2 & 22.0   & 84.36 & \textbf{89.20} \\
\rowcolor{lightblue}
+CAPL (Ours) & \textbf{86.20} & \textbf{6.5} & \textbf{18.4} & \textbf{84.62} & 88.49 \\

\bottomrule
\end{tabularx}

\end{table*}

\textbf{Results on Single-Image Tasks.}
As shown in the Tab.~\ref{tab:single_results}, although CAPL is primarily trained on multi-image samples and does not directly involve single-image task data, it still maintains stable performance across several single-image visual benchmarks, with improvements on some of them. For example, on Qwen2.5-VL, the POPE score increases from 81.23 to 82.94, while CHAIRs decreases from 22 to 18.4 on GLM4.1VBase. We conjecture that during training, CAPL effectively suppresses the model’s latent hallucination tendencies by pushing it away from negative samples, while preference learning over multi-image visual information enriches the model’s visual knowledge. As a result, the model can use visual signals more accurately even when reasoning over a single image.

\begin{table}[tb]
  \caption{Ablation study on our framework components: we evaluate results by progressively adding our cross-image attention and attentive preference training.}
  \label{tab:ablation_attention}
  \centering
  \small
  \begin{tabular}{lcccccc}
    \toprule
    \multirow{2}{*}{}
    & \multicolumn{2}{c}{Qwen2.5-VL} 
    & \multicolumn{2}{c}{InternVL2.5} 
    & \multicolumn{2}{c}{GLM4.1VBase} \\
    \cmidrule(lr){2-3} \cmidrule(lr){4-5} \cmidrule(lr){6-7}
    & BLINK & MUIRBench 
    & BLINK & MUIRBench 
    & BLINK & MUIRBench \\
    \midrule
    Base  & 54.60 & 58.42 & 54.81 & 48.54 & 58.17 & 57.84 \\
    +Attn & 55.34 & 58.96 & 55.02 & 49.07 & 58.23 & 58.07 \\
    \rowcolor{lightblue}
    Ours  & 57.76 & 62.00 & 55.76 & 52.12 & 61.33 & 60.57 \\
    \bottomrule
  \end{tabular}
\end{table}

\begin{table*}[htbp]
\caption{Ablation on negative sample construction across all subsets of MUIRBench. We compare the Base Model of GLM4.1VBase, DPO with original negatives, and DPO with truncated negatives.}
\centering
\label{tab:ablation_dpo}
\small
\begin{tabularx}{\textwidth}{l*{13}{>{\centering\arraybackslash}X}}
\toprule
Model & \rotatebox{90}{Overall} 
& \rotatebox{90}{Action}
& \rotatebox{90}{Similarity}
& \rotatebox{90}{Cartoon}
& \rotatebox{90}{Counting}
& \rotatebox{90}{Diagram}
& \rotatebox{90}{Difference}
& \rotatebox{90}{Geographic}
& \rotatebox{90}{I-T Match}
& \rotatebox{90}{Ordering}
& \rotatebox{90}{Scene}
& \rotatebox{90}{Grounding}
& \rotatebox{90}{Retrieval} \\
\midrule
GLM4.1VBase
& 57.9 & 46.3 & 58.7 & 46.2 & 39.3 & 73.9 & 57.9 & 41.0 & 68.8 & 20.3 & 66.1 & 28.6 & 59.6 \\

+Attn 
& 58.1 & 47.0 & 56.6 & 46.2 & 40.2 & 74.9 & 57.4 & 41.0 & 68.8 & 23.4 & 65.6 & 28.6 & 61.0 \\

$\oplus$ Original DPO 
& 59.5 & 46.3 & 62.8 & 47.4 & 40.6 & 75.6 & 58.5 & 44.0 & 73.3 & 20.3 & 69.4 & 31.0 & 55.8 \\

\rowcolor{lightblue}
$\oplus$ Truncated DPO 
& 60.6 & 48.2 & 59.7 & 48.7 & 42.7 & 76.9 & 60.3 & 48.0 & 72.6 & 21.9 & 70.4 & 31.0 & 59.6 \\

\midrule
$\Delta$ 
& +2.7 
& +1.8 
& +1.0 
& +2.6 
& +3.4 
& +3.0 
& +2.4 
& +7.0 
& +3.9 
& +1.6 
& +4.3 
& +2.4 
& +0.0 \\
\bottomrule
\end{tabularx}
\end{table*}
\subsection{Ablation Studies}
\textbf{The Effect of Cross-Image Attention and Training Framework.}
From Tab.~\ref{tab:ablation_attention}, we observe that introducing selective cross-image attention (+Attn) consistently yields modest yet stable improvements across all three backbone models on both BLINK and MUIRBench. This indicates that enhancing cross-image information interaction at inference time alone can partially alleviate erroneous associations in cross-image reasoning, although the overall gains remain limited.

When further combined with our proposed preference optimization training, the performance improvements become substantially more pronounced, particularly on MUIRBench (\textit{e.g.}, Qwen2.5-VL improves to 62.00 and GLM4.1VBase reaches 60.57). This observation suggests a synergistic effect between structured cross-image modeling and targeted preference optimization: the former explicitly strengthens the model’s ability to capture inter-image dependencies, while the latter leverages preference signals to further regularize the generation process, thereby more effectively mitigating multi-image hallucination. 

\setlength{\intextsep}{2pt}
\setlength{\abovecaptionskip}{0pt}
\setlength{\belowcaptionskip}{2pt}
\begin{wrapfigure}{l}{0.54\linewidth}
    \centering
    \includegraphics[width=0.5\textwidth]{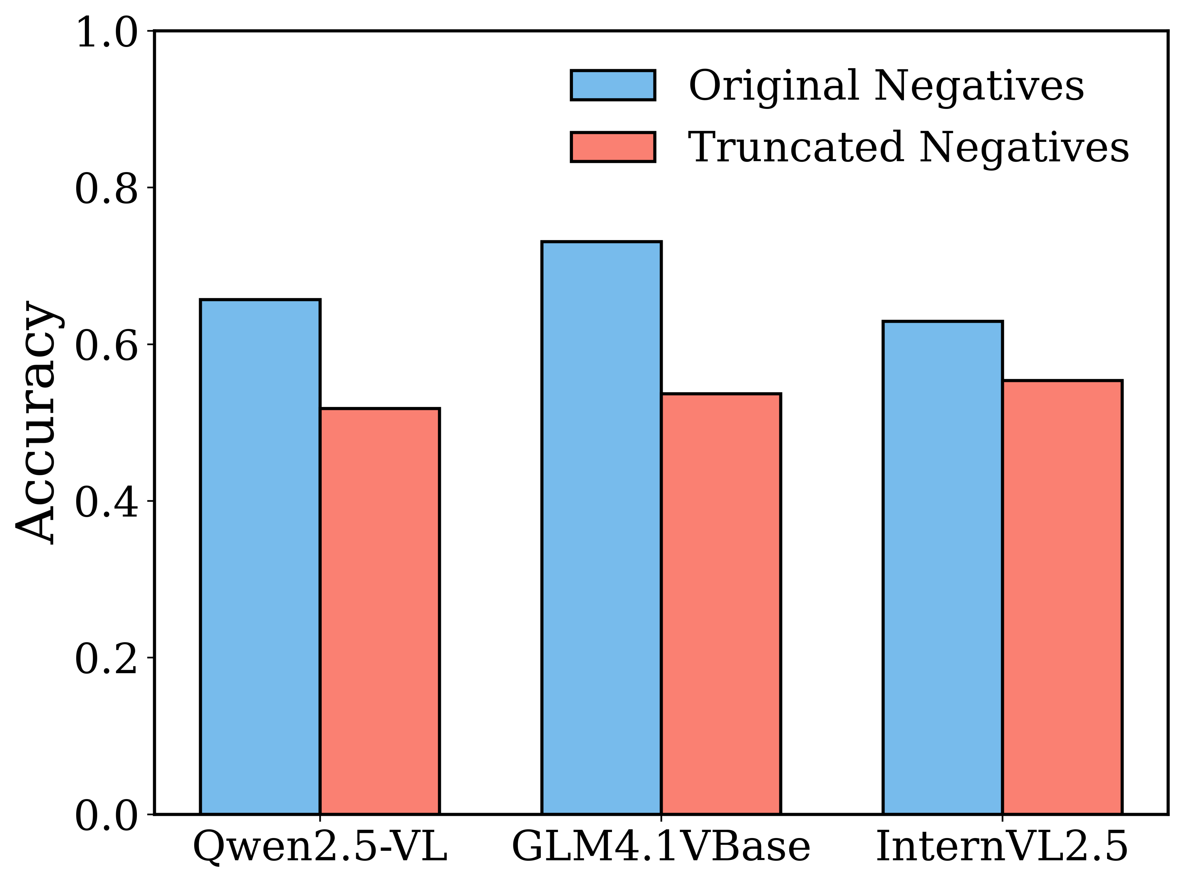}
    \caption{Comparison of three models’ accuracies on two types of negative samples: those generated by the original model structure (Original Negatives) and those generated using the truncated attention structure (Truncated Negatives). Lower accuracy indicates that the negative samples are more challenging.}
    \label{fig:ablation_negative}
\end{wrapfigure}

\textbf{The Effect of Truncated Attention–Induced Negatives}
In DPO training, the quality of negative samples is crucial, as samples that deviate further from the true distribution typically provide stronger optimization signals. To verify the effectiveness of Truncated Attention–Induced Negatives, we first compared the accuracy of negative samples generated by the original model with those generated using truncated attention. The results in Fig.~\ref{fig:ablation_negative} show that the induced negatives exhibit consistently lower accuracy, decreasing by about 20\% on GLM4.1VBase. This result indicates that truncating cross-image attention effectively exposes more challenging error cases. Building on this observation, we conducted comparative experiments across all MUIRBench subtasks using GLM4.1VBase, including the Base model, +Attention, original DPO, and Truncated DPO. The results in Tab.~\ref{tab:ablation_dpo} demonstrate that Truncated DPO performs particularly well on representative tasks, such as Geographic Understanding (+7.0), Scene Understanding (+4.3), and Difference Spotting (+2.4), outperforming the original DPO on most subtasks. By restricting cross-image information flow during negative sample generation, truncated attention constructs more challenging erroneous samples, resulting in a clearer preference gap during DPO training. Additionally, we designed some negative samples whose final answers are correct, but due to blocked inter-image attention, the reasoning paths are incorrect; these are also used in DPO to help the model learn to avoid potential errors.

\begin{figure}[tb]
  \centering
  \includegraphics[width=\linewidth]{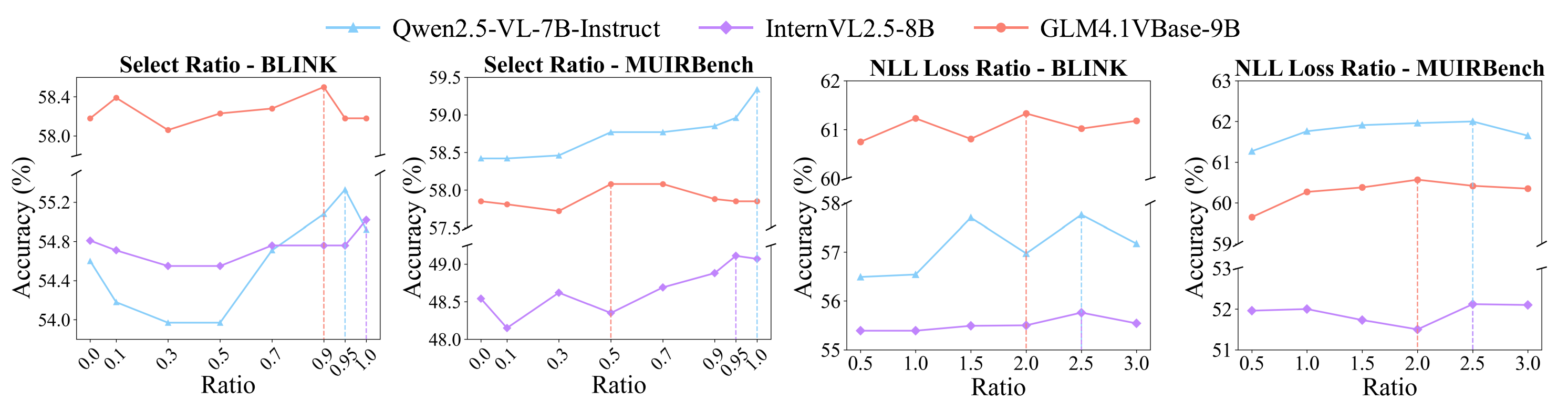}
      \caption{Ablation results with respect to the Select Ratio $\rho$ and the NLL Loss Ratio $\lambda$. }
  \label{fig:ablation}
\end{figure}

\textbf{Ablation on the Select Ratio $\rho$.}
The selection ratio $\rho$ is a key parameter in cross-image attention, controlling the number of tokens involved in cross-image interactions. When $\rho$ is too small, information flow is restricted and cross-image alignment becomes weaker. conversely, an excessively large $\rho$ introduces redundant noise and amplifies interference. The results in Fig.~\ref{fig:ablation}(Left) show that performance improves as $\rho$ increases but declines slightly when it approaches 1, indicating that a small portion of image tokens may contain noise. Retaining most, but not all, key tokens for cross-image interaction achieves a better balance between effective modeling and noise suppression. Empirically, the optimal values are $\rho=0.95$ for Qwen2.5-VL and InternVL2.5, and $\rho=0.9$ for GLM4.1VBase.

\textbf{Ablation on the NLL Loss Ratio $\lambda$.}
The NLL loss ratio $\lambda$ determines the weight of the NLL loss in the overall training objective. When $\lambda$ is too small, the model relies excessively on the DPO signal, which may weaken its language modeling capability, while a overly large $\lambda$ diminishes the preference distinction signal and may affect optimization stability. Experimental results in Fig.~\ref{fig:ablation}(Right) show that, considering performance on multi-image hallucination benchmarks BLINK and MUIRBench, Qwen2.5-VL and InternVL2.5 achieve optimal performance at $\lambda=2.5$, whereas GLM4.1VBase performs best at $\lambda=2$.

\section{Conclusion}
This work proposes \textbf{C}ross-Image \textbf{A}ttention calibration and \textbf{P}reference \textbf{L}earning \textbf{(CAPL)}, a framework designed for multi-image tasks. It enhances inter-image information flow through selective cross-image attention and introduces a preference optimization strategy based on a contrast of reasoning results between full cross-image interaction and truncated images, encouraging predictions grounded in true visual evidence. Experimental results show that each component of our method contributes to performance improvements, and the combined framework significantly enhances model performance on multi-image hallucination and general tasks, while maintaining single-image reasoning capabilities, demonstrating the practicality and effectiveness of cross-image attention and attentive preference learning design in complex multi-image scenarios.

\clearpage  


%
%
\bibliographystyle{splncs04}
\bibliography{main}
\end{document}